\documentclass{article}

\usepackage{arxiv}
\usepackage{array}
\usepackage[utf8]{inputenc} 
\usepackage[T1]{fontenc}    
\usepackage{hyperref}       
\usepackage{url}            
\usepackage{booktabs}       
\usepackage{amsfonts}       
\usepackage{nicefrac}       
\usepackage{microtype}      
\usepackage{lipsum}		
\usepackage{graphicx}
\usepackage[square,numbers]{natbib}
\usepackage{multirow}
\usepackage{doi}
\usepackage{nomencl}

\title{Data-Driven Design-by-Analogy:\\ 
State of the Art and Future Directions}

\date{June 1, 2021}	

\author{ 
{
\hspace{1mm}Shuo Jiang}\thanks{Contact email: \texttt{jsmech@sjtu.edu.cn}} \\
	Shanghai Jiao Tong University\\
	800 Dongchuan Road, Shanghai, China, 200240\\
	\texttt{jsmech@sjtu.edu.cn} \\
	
	\And
	{\hspace{1mm}Jie Hu} \\
	Shanghai Jiao Tong University\\
	800 Dongchuan Road, Shanghai, China, 200240\\
	\texttt{hujie@sjtu.edu.cn} \\
	
	\And
	{\hspace{1mm}Kristin L. Wood} \\
	University of Colorado Denver\\
	1201 Larimer St, Denver, CO, USA, 80204\\
	\texttt{kristin.wood@ucdenver.edu} \\
	
	\And
	{\hspace{1mm}Jianxi Luo} \\
	Singapore University of Technology and Design\\
	8 Somapah Road, Singapore, 487372\\
	\texttt{luo@sutd.edu.sg} \\
	
}

\renewcommand{\headeright}{Data-Driven Design-by-Analogy: State of the Art and Future Directions}
\renewcommand{\undertitle}{}
\renewcommand{\shorttitle}{}

\hypersetup{
pdftitle={A template for the arxiv style},
pdfsubject={q-bio.NC, q-bio.QM},
pdfauthor={David S.~Hippocampus, Elias D.~Striatum},
pdfkeywords={First keyword, Second keyword, More},
}

\begin{document}
\maketitle

\begin{abstract}
	Design-by-Analogy (DbA) is a design methodology wherein new solutions, opportunities or designs are generated in a target domain based on inspiration drawn from a source domain; it can benefit designers in mitigating design fixation and improving design ideation outcomes. Recently, the increasingly available design databases and rapidly advancing data science and artificial intelligence technologies have presented new opportunities for developing data-driven methods and tools for DbA support. In this study, we survey existing data-driven DbA studies and categorize individual studies according to the data, methods, and applications in four categories, namely, analogy encoding, retrieval, mapping, and evaluation. Based on both nuanced organic review and structured analysis, this paper elucidates the state of the art of data-driven DbA research to date and benchmarks it with the frontier of data science and AI research to identify promising research opportunities and directions for the field. Finally, we propose a future conceptual data-driven DbA system that integrates all propositions.
\end{abstract}


\makenomenclature
\nomenclature{\texttt{DbA}}{Design-by-Analogy}
\nomenclature{\texttt{LISA}}{Learning and Inference with Schemas and Analogies}
\nomenclature{\texttt{AMBR}}{Associative Memory-Based Reasoning}
\nomenclature{\texttt{ARGO}}{A system for Design-by-Analogy}
\nomenclature{\texttt{KRITIK}}{An early case-based design system}
\nomenclature{\texttt{IDEAL}}{Integrated "Design by Analogy and Learning” system}
\nomenclature{\texttt{IEKG}}{Interdisciplinary Engineering Knowledge Genome}
\nomenclature{\texttt{DANE}}{Design by Analogy to Nature Engine }
\nomenclature{\texttt{PAnDA}}{Product Aspects in Design by Analogy}
\nomenclature{\texttt{SEABIRD}}{Scalable search for systematic biologically inspired design}
\nomenclature{\texttt{D-APPS}}{Design Analogy Performance Parameter System}
\nomenclature{\texttt{DRACULA}}{Design Repository \& Analogy Computation via Unit Language Analysis}
\nomenclature{\texttt{InnoGPS}}{Innovation Global Positioning System}
\nomenclature{\texttt{TechNet}}{Technology Semantic Network}
\nomenclature{\texttt{SOLVENT}}{A mixed initiative system for finding analogies between research papers}
\nomenclature{\texttt{VISION}}{Visual Interaction tool for Seeking Inspiration based On Nonnegative Matrix Factorization}
\nomenclature{\texttt{FOBIE}}{Focused Open Biology Information Extraction}
\nomenclature{\texttt{FBS}}{Function-Behavior-Structure model}
\nomenclature{\texttt{E2B}}{Engineering-to-Biology}
\nomenclature{\texttt{BID}}{Bio-Inspired Design}
\nomenclature{\texttt{AI}}{Artificial Intelligence}
\nomenclature{\texttt{XAI}}{Explainable Artificial Intelligence}
\nomenclature{\texttt{NLP}}{Natural Language Processing}
\nomenclature{\texttt{PCA}}{Principal Component Analysis}
\nomenclature{\texttt{ANN}}{Artificial Neural Network}
\nomenclature{\texttt{RNN}}{Recurrent Neural Network}
\nomenclature{\texttt{Bi-RNN}}{Bi-directional Recurrent Neural Network}
\nomenclature{\texttt{CNN}}{Convolutional Neural Network}
\nomenclature{\texttt{GNN}}{Graph Neural Network}
\nomenclature{\texttt{GAN}}{Generative Adversarial Network}
\nomenclature{\texttt{VAE}}{Variational Autoencoder }
\nomenclature{\texttt{LSTM}}{Long short-term memory network}
\nomenclature{\texttt{AHC}}{Agglomerative Hierarchical Clustering}
\nomenclature{\texttt{RSC}}{Relevance Score-based Clustering}
\nomenclature{\texttt{SVM}}{Support Vector Machine}
\nomenclature{\texttt{NB}}{Naive Bayes}
\nomenclature{\texttt{KNN}}{K-Nearest Neighbors}
\nomenclature{\texttt{LSA}}{Latent Semantic Analysis}
\nomenclature{\texttt{LDA}}{Latent Dirichlet Allocation}
\nomenclature{\texttt{NMF}}{Nonnegative Matrix Factorization}
\nomenclature{\texttt{VSM}}{Vector Space Method}
\nomenclature{\texttt{USPTO}}{United States Patent and Trademark Office}

\printnomenclature

\section{Introduction}
\label{sec1}
Design-by-Analogy (DbA) is a design methodology, wherein new solutions are generated in a target domain based on inspiration drawn from a source domain through cross-domain analogical reasoning \cite{ullman1992mechanical,goel1997design,christensen2007relationship}. DbA is an active research area in engineering design and various methods and tools have been proposed to support the implement of its process \cite{linsey2012design, murphy2014function,Song2019DbA,Song2020Anexp,Goel2015Analogical}. Studies have shown that DbA can help designers mitigate design fixation \cite{linsey2010study} and improve design ideation outcomes \cite{fu2013meaning}.

\begin{figure}
	\centering
	\includegraphics[width=8cm]{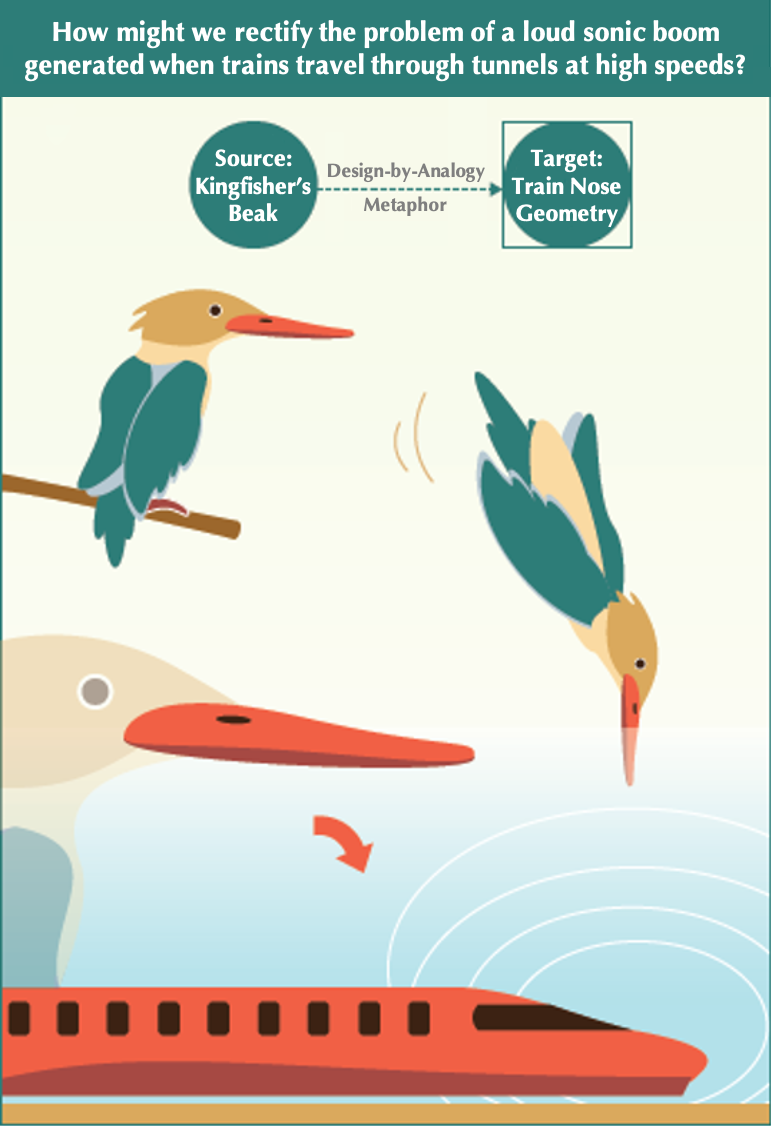}
	\caption{Exemplar of DbA: high speed train nose design inspired by kingfisher’s beak \cite{Lauff2021}}
	\label{fig:fig1}
\end{figure}

Fig.\ref{fig:fig1} presents an example of DbA applications \cite{Lauff2021}. This case aims to solve an engineering design problem: How might we rectify the loud sonic boom generated when trains travel at high speeds through tunnels in atmospheric conditions \cite{Lauff2021,linic2020experimental}? For potential design solutions to this problem, engineers explored structures in other design fields than trains or in the nature that effectively “break” the sonic-boom effect. When looking into the nature, engineers discovered that kingfisher birds could slice through the air and dive into the water at extremely high speeds to catch prey while barely making a splash. By analogy, engineers re-designed the train's front-end nose to mimic the geometry of the kingfisher’s beak. This analogical design reduced noise and eliminated tunnel booms. It also allowed the train to travel 10\% faster consuming 15\% less energy\footnote{https://asknature.org/innovation/high-speed-train-inspired-by-the-kingfisher/}. The example shows a practical case where design-by-analogy leads to creative problem solving and dramatic performance improvement.

In recent years, the increasingly available design databases and rapidly advancing data science and artificial intelligence (AI) technologies have enabled new data-driven methods and tools that support DbA. For example, deep learning, knowledge graph, natural language understanding, and computer vision may support the analogy representation, retrieval, mapping, and evaluation processes \cite{Jiang2021,Sarica2021a,Luo2021,kruiper-etal-2020-laymans,han2018combinator,han2018computational,han2020data}. To the best of our knowledge, there is no systematic review and structured analysis of the literature on data-driven DbA to date. Therefore, to reveal promising research opportunities and future directions, we reviewed the state-of-the-art data-driven methods and tools for DbA and benchmarked them with the latest data science and AI technologies.

This paper proceeds as follows. Section 2 provides the theoretical background of DbA, including the analogical thinking models in cognitive psychology and experimental understanding of the effect of DbA on creativity in the engineering design field. Section 3 describes the research methodology. Section 4 presents a nuanced organic review of the data-driven DbA research literature individually. Section 5 provides a structured and integrated analysis of all the literature together, from three perspectives: data, methods, and applications, to reveal the general state-of-the-art of the field. Section 6 maps the course of feasible directions for future research from three aspects and their interactions. Finally, section 7 concludes the paper.

\section{Background}
\label{sec:sec2}
Cognitive psychology reveals that people often tend to solve given problems by mapping solutions from known questions \cite{ross1987like,markman1997,markman1993structural,kokinov2003computational,hall1989computational,french1985conceptual}, i.e., analogical reasoning. Markman \cite{markman1997,markman2009supporting} describes analogy as a powerful cognitive mechanism that allows two objects to be seen as similar based on the connected systems of relations between them. Making analogical inferences can improve critical thinking and logical reasoning \cite{markman1993structural}. Gentner et al. \cite{Gentner2013} suggested that making analogy is a key ability that distinguishes humans from other species and introduced a three-phase model of the analogical reasoning process including retrieval, mapping, and evaluation. Although other researchers proposed slightly different subdivisions, they broadly agreed on these phases \cite{markman1993structural,kokinov2003computational,hall1989computational}. Human analogical processes generally begin with storing and encoding the source knowledge (analogy candidates) of memory, followed by recognizing and searching the source analogy based on the given target problem. The source analogy then serves as the base model through which structured knowledge can be understood, applied, or projected on to a less familiar or abstract domain \cite{ward2010cognition,hey2008analogies}. In the retrieval and mapping processes, three types of similarities are usually exploited, namely, literal, attributional, and relational similarities \cite{Verhaegen2011}. Prior studies have revealed that a good analogy often involves an alignment of relational structure \cite{gentner1997structure}. After mapping, the final step, i.e., evaluation is performed, which entails judging the analogy along with any generated inferences. In addition to the above-mentioned steps, other subphases, such as abstraction and parallel subprocess interaction, have also been described as essential cognitive processes in analogical reasoning \cite{chalmers1992high}. 

Since the previous century, several computation-based models of analogy-making have been proposed \cite{evans1964program,hummel1997distributed,kokinov2000dynamic,Grace2015}. French \cite{French2002} classified these models into three broad groups based on their underlying architecture. (1) Symbolic models are largely part of the symbolic paradigm in the AI field, in which symbols, logic, search, planning, and means–ends analysis, play an essential role. One of the most famous symbolic models is \textbf{ANALOGY} \cite{evans1964program}, proposed by Evans, which was designed to conduct proportional analogies of the form “A:B::C:?” obtained from the test questions of geometric figures. (2) Connectionist models adopt the framework of complex networks, including nodes, edges, and their weights. For example, the \textbf{LISA} analogy engine \cite{hummel1997distributed} relies on partially distributed representations of concepts, selective activation, and dynamic binding to associate the relevant structures to support both analogy-retrieval and mapping. (3) Hybrid models share the features of both connectionist and symbolic models, and they typically comprise neurons that enable symbolic interpretation or interactions. For instance, the \textbf{AMBR} model \cite{kokinov2000dynamic} consists of many micro-agents, each of which represents a small piece of knowledge that enables analogy-making.

Using analogical reasoning to generate new design solutions or concepts is referred as Design-by-Analogy in the field of engineering design. Gill et al. \cite{gill2019dimensions} and Tsoka et al. \cite{tsoka2020dimensions} investigated the dimensions of product similarity that designers tend to use to identify the source product when using DbA in engineering design. These empirical studies indicate designers utilize at least six different dimensions to draw analogy between target and source products: working principle, structure, human interaction, function, energy flows, and material flows. Among them, working principle, structure, and human interaction are more dominant in driving analogy than the other dimensions. In addition, engineering design researchers have examined how the analogical distance \cite{fu2013meaning,srinivasan2018does,song2017patent,chan2011benefits,chiu2012investigating,malaga2000effect,enkel2010creative}, representation of modality \cite{chan2011benefits,linsey2008modality,atilola2015}, commonness \cite{chan2011benefits}, amount or quantity \cite{Song2018Char}, time of involvement of analogy \cite{Tseng2008}, and type of similarity \cite{fu2015design,moreno2016overcoming} influence the solutions or concepts generated using the DbA. Among these factors, the influence of the analogical distance between the source and target domains has received the most attention. 

Recent researchers have advanced the understanding of the multifaceted effects of the analogical distance on ideation behaviors and outcomes. The conceptual leap hypothesis states that distant analogies give rise to novelty and breakthroughs with lower efficacy, owing to surface dissimilarities \cite{ward1998analogical}. This statement was corroborated by several empirical results from human-based studies. For example, according to Chan et al. \cite{chan2011benefits}, far-field stimuli yields more novel but fewer ideas than the near-field stimuli. Srinivasan et al. \cite{srinivasan2018does} stated that with the increase in the analogical distance, the novelty of the newly generated ideas increases, while both the quantity and quality decrease. Within the different experimental and empirical contexts, Fu et al. \cite{fu2013meaning} observed that a moderate analogical distance might provide trade-offs and yield optimal design ideation performance. 

Despite its usefulness, the efficacy of DbA is naturally limited by the designer’s prior knowledge and memory for potential retrieval and mapping to a design problem. This challenge is greater for inexperienced or specialized designers with a limited scope of knowledge. Meanwhile, not every designer is proficient in analogical thinking. In fact, these conditions suggest the fundamental values of using external knowledge database as the source and using intelligent algorithms to carry out analogical reasoning. Then, the nuanced experimental findings on the analogical distance and other aspects, as reviewed in this section, may guide the retrieval of design analogies from a large-scale design database and the analogical reasoning process in data-driven workflows.

\section{Research Methodology}
\label{sec:sec3}
This paper aims to elucidate the state-of-the-art of data-driven DbA research to date, especially when it comes to the application of AI and data science technologies. An advanced search in the Web-of-Science database was conducted to generate an initial list of data-driven DbA publications\footnote{Literature retrieval date: Feb 18, 2021.}. With this search query \emph{“TS = ((Design-by-Analogy) AND data) AND PY=1950-2021”}\footnote{“TS” stands for “topic” and “PY” stands for “published year” in the Web-of-Science search query.}, we identified 23 relevant papers. 

We manually examined the content of these publications and filtered out 4 articles that did not fall within the ambit of the engineering or innovation field and 6 articles that do not present data-driven methods or tools for DbA. Among the remaining ones, 5 of them are short version conference papers of the full version journal papers and are also excluded. Meanwhile, we found 39 additional relevant studies on data-driven DbA methods or tools by a snowballing search through the backward and forward citations of all the papers in the initial set (including those excluded ones in the previous step). For example, we were able to identify some recent efforts on data-driven DbA support tools, such as Idea-Inspire 4.0 \cite{siddharth2018evaluating}, TechNet \cite{sarica2020technet}, and B-link \cite{Shi2017}, from the forward citations of a paper titled \emph{“Function-based \textbf{\underline{design-by-analogy}}: a functional vector approach to analogical search”} \cite{murphy2014function}. From the backward citation of another paper titled \emph{“Identifying candidates for \textbf{\underline{design-by-analogy}}”} \cite{Verhaegen2011}, we could identify the first generation of DbA systems, such as KRITIK \cite{goel1997kritik} and IDEAL \cite{bhatta1996design}. Such newly identified studies directly refer to design-by-analogy or its synonym in the body text or discuss its usage to support one or more DbA subprocesses. 

\begin{figure}
	\centering
	\includegraphics[width=14cm]{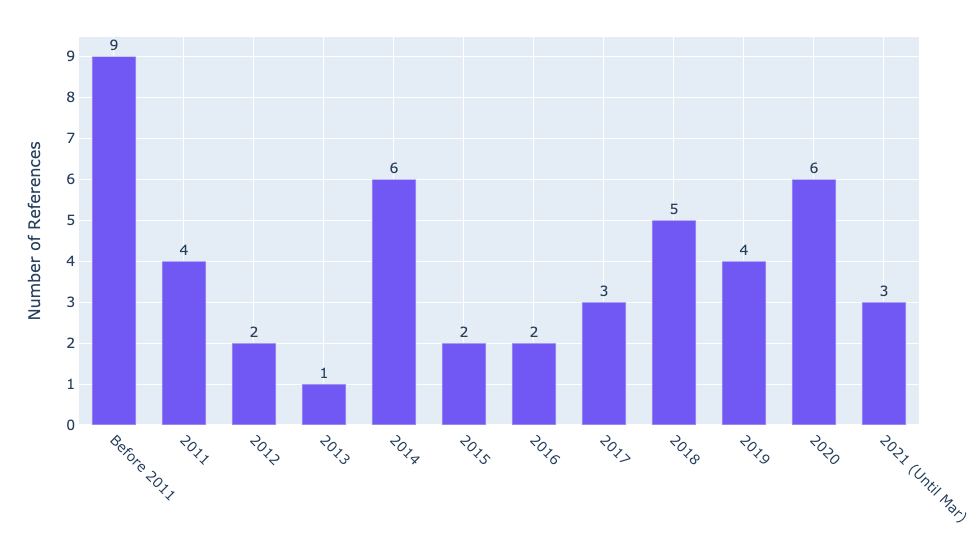}
	\caption{Number of data-driven DbA publications per year }
	\label{fig:fig2}
\end{figure}

\begin{figure}
	\centering
	\includegraphics[width=13cm]{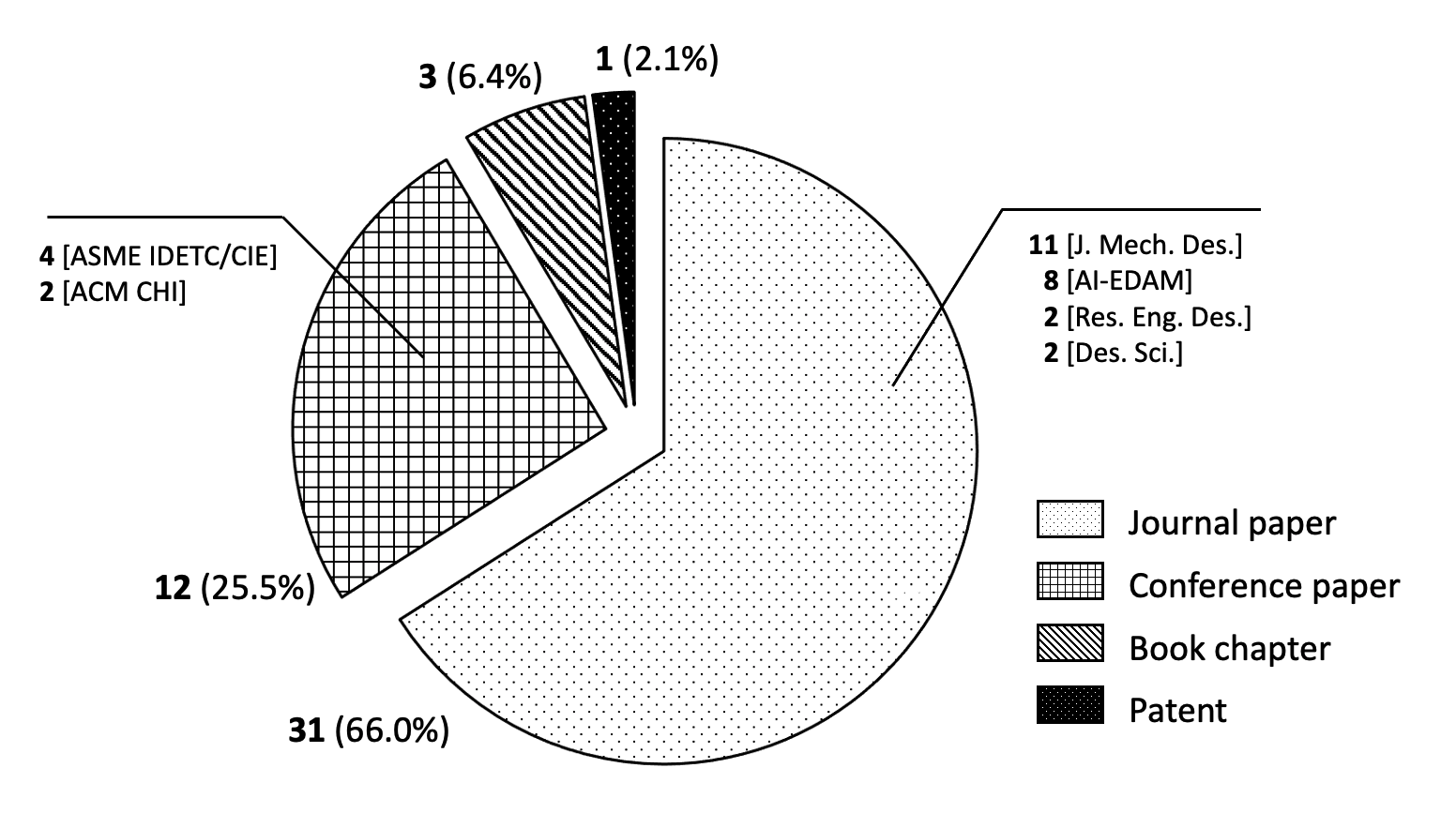}
	\caption{Distribution of data-driven DbA publications by literature types}
	\label{fig:fig3}
\end{figure}

After these steps, the final literature set includes 47 papers\footnote{The final literature set includes these references: [4-7,13,15–19,30,41,50,53–60,62–72,77–89,91,92].}. As observed in Fig.~\ref{fig:fig2}, the number of publications exhibited an upward trend in recent years, indicating a growing interest in this field. Fig.~\ref{fig:fig3} shows the proportional breakdown of publications by types, including journal articles, conference papers, book chapters, and patents. In Fig.~\ref{fig:fig3}, we marked out four journals and two conferences that include more than one papers in the final set:

Journals - (1) Journal of Mechanical Design, (2) Artificial Intelligence for Engineering Design, Analysis and Manufacturing, (3) Research in Engineering Design, and (4) Design Science.

Conferences - (1) International Design Engineering Technical Conferences \& Computers and Information in Engineering Conference (IDETC/CIE) held by ASME and (2) Conference on Human Factors in Computing Systems (CHI) held by ACM.

\section{Data-driven DbA methods and tools: an organic review}
\label{sec:sec4}
The first generation of DbA systems, most of which were knowledge-based expert systems, were developed before the 2000s. Huhns and Acosta \cite{huhns1988argo} developed the \textbf{ARGO}, which represents problem-solving plans using rule-based graphs that enable analogical retrieval for solving new problems. Goel et al. \cite{goel1997kritik} proposed KRITIK, which represents design cases through function-behavior-structure (FBS) models \cite{qian1996function}. KRITIK retrieves specific designs based on the functional information of query design, suggests modifications, and allows the verification of the newly generated designs. The same research group developed the \textbf{IDEAL} \cite{bhatta1996design}, which is also based on the FBS models. The IDEAL system allows the extraction of generic teleological mechanisms and performs analogical mapping by comparing two designs with different patterns. 

With the rise of data science in the past two decades, various data-driven methods and tools have been developed to support DbA. Reich and Shai \cite{reich2012interdisciplinary} proposed \textbf{the infused design method} that represents design problems at the meta-level, to facilitate the discovery and use of knowledge across different technological domains. Based on the infused design theory, the interdisciplinary engineering knowledge genome (\textbf{IEKG}) was developed to support the search for general knowledge and method structures in the interdisciplinary domains \cite{reich2012interdisciplinary}. Verhaegen et al. \cite{Verhaegen2011} distilled product features using the word co-occurrence analysis and principal component analysis (PCA), to automatically and systematically retrieve candidate products for DbA. Linsey et al. \cite{linsey2012design} proposed the \textbf{WordTree} method to linguistically re-represent design problems based on a predefined large-scale semantic network (WordNet) that can help designers seek potential analogies for idea generation. Murphy et al. \cite{murphy2014function} proposed a \textbf{functional vector method} to represent design documents into vectors based on the word frequency of functional verbs from the Functional Basis model \cite{stone2000development}. A further investigation has revealed that functionally novel concepts can be generated using the functional analogy search method \cite{fu2015design}. 

Sanaei et al. \cite{Sanaei2017} developed an analogical retrieval method, whereby text mining and a recurrent neural network (RNN)-based ensemble model are used to extract relational structure from the text. Gilon et al. \cite{gilon2018analogy} proposed an analogical search engine to search distant but relevant design stimuli for specific needs using the information extraction strategy and bidirectional RNN deep learning model. Chan et al. \cite{Chan2018} developed a mixed initiative system named \textbf{SOLVENT} to identify analogies among scientific papers. The system annotates four aspects of research papers (background, purpose, mechanism, and findings), and utilizes the Doc2Vec pre-trained word embeddings to represent and map different contents. Han et al. \cite{han2018computational,han2018combinator} developed a series of computer-based tools to support DbA, including the \textbf{retriever} and \textbf{combinator}. The retriever takes ConceptNet as the knowledge base to construct the base ontology for a given design problem and search for less familiar target ontology for design ideation \cite{han2018computational}. The combinator generates combinational prompts in both text and image forms by combining indirectly related ideas based on a customized design knowledge database constructed using ConceptNet through web crawling and text mining \cite{han2018combinator}. 

Among the various types of engineering design-related data sources, the patent database has attracted significant scholarly attention because of its large size, diversity, and rich design-related information in its documents. Altshuller \cite{al1999} extracted various heuristic design rules, or principles, by analyzing abstract patterns in the global patent database and developed a problem-solving framework named TRIZ to guide engineers to overcome impasses in analogical reasoning. Subsequently, various data-driven tools have been proposed to support the use of TRIZ. Cascini and Russo \cite{Cascini2007} developed a system named \textbf{PAT-ANALYZER} to automatically construct design contradictions in TRIZ by mining patent textual information. Vincent et al. \cite{vincent2002systematic} proposed \textbf{Bio-TRIZ}, which expands on the biological information and principals in the TRIZ system. Beyond TRIZ-related tools, McCaffrey and Spector \cite{Mccaffrey2016} developed \textbf{Analogy Finder}, a DbA support system to identify adaptable analogies in the patent database by rephrasing the description of the design problem as verbs and their synonyms. Luo et al. \cite{Luo2021} developed \textbf{InnoGPS}, a computer-aided ideation tool based on the total technology space constructed in the entire United States Patent and Trademark Office (USPTO) database. The InnoGPS system can guide designers to search and retrieve knowledge and concepts for drawing analogies across domains based on knowledge distance. Similarly, Song et al. \cite{song2017patent} utilized the technological network to locate identified patent analogies in home-, near-, and far-field stimuli based on the clustering partition results. 

Recently engineering design researchers have used the latest natural language processing (NLP) and text mining techniques to construct semantic networks from engineering design-related databases, such as \textbf{B-link} \cite{Shi2017} based on approximately one million engineering papers, and \textbf{TechNet} \cite{sarica2020technet} based on six million patent documents. Both B-Link and TechNet have been used for analogical retrieval \cite{Shi2017,sarica2019engineering}, concept generation \cite{Sarica2021idea,chen2019}, and evaluation \cite{han2020data}. In addition, several topic modeling algorithms have been used to structure and visualize textual design database for exploration and inspiration. Fu et al. \cite{fu2013discovering} utilized the combination of a Bayesian model and latent semantic analysis to map patent documents in a network structure; this method provides a basis for automated retrieval of cross-domain inspirational analogy. The analogical distance between the initial design problem and potential stimuli is defined as the path length in a Bayesian network. Song et al. \cite{Song2019DbA} applied the hierarchical nonnegative matrix factorization (NMF), a topic modeling algorithm, to structurally represent patent database with three facets: behavior, material, and component. Based on this clustered structure of patent data, they developed \textbf{VISION} \cite{Song2020Anexp}, a visual interaction tool for seeking analogical inspiration and data exploration.

Bio-inspired design (BID) \cite{vincent2006biomimetics,chakrabarti2010biologically,nagel2012art,fu2014bioinspired} is a form of DbA whereby inspiration is drawn from natural biological phenomenon for engineering design. In recent years, many efforts have been made to develop data-driven methods, tools, and databases for BID. Cheong et al. \cite{cheong2011bio} and Nagel et al. \cite{nagel2010engineering} developed the \textbf{engineering-to-biology thesaurus} by mining meaningful keywords from biological text aligned with the engineering functional terms in the functional basis. The thesaurus serves as the basis for engineers to find biological analogies and identify the functional reasoning linking two domains when solving design problems. Goel et al. \cite{vattam2011dane} developed the Design by Analogy to Nature Engine (\textbf{DANE}), a design case library containing approximately 40 FBS models of natural and artificial systems. DANE also provides a framework for users to retrieve previous cases based on queries in multiple forms. Helms and Goel \cite{helms2014thefour} proposed the \textbf{four-box method} to manually represent design problems in four facets (functions, environments, specifications, and performance) based on FBS models and evaluate analogies using the heuristic T-chart in the BID. Cheong and Shu \cite{cheong2014retrieving} proposed the syntactic rule-based computational method to identify biological analogies that involve causal relations. Glier et al. \cite{giler2014exploring} explored the use of automated text classifiers for BID, including naive Bayes classifiers, k-nearest neighbors, and support vector machine, to determine whether a textual stimuli is helpful for solving the given design problem. The Biomimicry 3.8 Institute \cite{deldin2014asknature} developed a web-based tool, \textbf{AskNature}, which employs a dedicated taxonomy to manage the more than 1,600 biological cases in its repository, to provide biological analogies for engineering designers. 

Verhaegen et al. \cite{verhaegen2011effectiveness} developed product aspects in DbA (\textbf{PAnDA}), a tool to identify candidate products for DbA by extracting the product aspects from texts through rule-based text-mining techniques. They also developed SEABIRD \cite{vandevenne2016seabird} based on the same strategy to map the technical systems in patent documents and biological systems in academic papers. \textbf{SEABIRD} enables the scalable search for biological stimuli for designers. Lucero et al. \cite{lucero2014,lucero2015design} integrated the functional basis model, WordNet, and the AskNature repository to develop the \textbf{D-APPS} tool and \textbf{DRACULA} software, which enable analogy matching between biological and engineering concepts and analogy searching for BID. Siddharth and Chakrabarti \cite{siddharth2018evaluating} developed the \textbf{Idea-Inspire 4.0} tool, which represents both biological and engineering concepts based on the domain-agnostic SAPPhIRE model ontology \cite{chakrabarti2005functional}. Idea-Inspire 4.0 supports the analogical retrieval and evaluation based on a manually created database of 60 biological concepts and 83 engineered concepts. More recently, Kruiper et al. \cite{kruiper-etal-2020-laymans,kruiper-etal-2020-scientific} built a dataset named focused open biology information extraction (\textbf{FOBIE}) and used it to train an NLP model to extract semi-open trade-off relations and arguments from scientific biological documents, providing a high-level filter for engineers. Bhasin et al. \cite{bhasin2021product} proposed a reduced function-means tree to distill both biological adaptations and their associated product architecture from existing bioinspired design abstraction database and product architecture representation database.

Although most of the aforementioned methods and tools entailed processing textual information, visual or multimodal analogy for design inspiration has been explored in recent research. Kwon et al. \cite{kwon2019visual} developed a framework to exploit visual similarity based on the web-based search engine to find visual analogies for further idea generation. Jiang et al. \cite{Jiang2021} proposed a convolutional neural network (CNN)-based model to automatically derive the vector space and design the feature vectors to represent patent images. The derived feature vectors that embed both visual information and technology-related knowledge can potentially facilitate the retrieval of visual stimuli based on the vector distances. Zhang et al. \cite{zhang2020unsupervised} proposed an unsupervised deep learning model, Sketch-pix2seq, which is trained using reconstruction and clustering losses to allow the extraction of the underlying shape features of sketches in the Quickdraw database. The constructed latent space for sketches provides a new way to define visual similarities and search the analogical sketches for DbA.

\section{Structure and state of the field: a synthesis}
\label{sec:sec5}
\subsection{Overview}
\begin{figure}
	\centering
	\includegraphics[width=15cm]{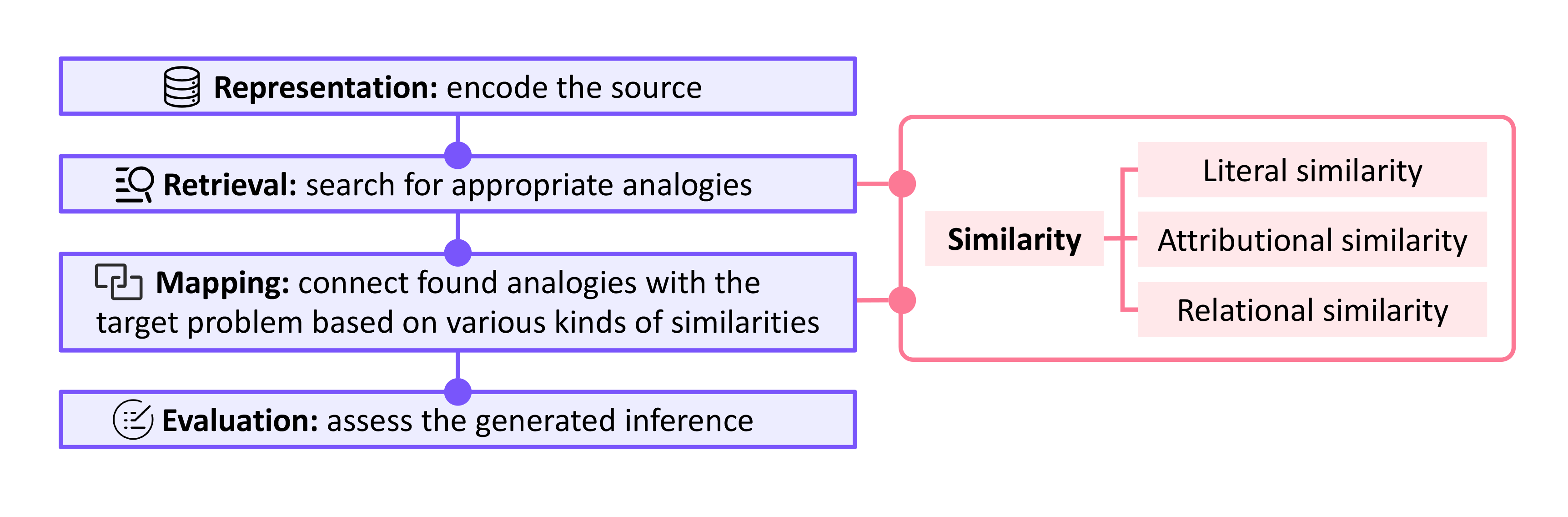}
	\caption{The four-phase model of making analogy in cognitive science}
	\label{fig:fig4}
\end{figure}

In section 4, we have presented a nuanced organic review of each data-driven DbA method or tool identified via our literature search. To elucidate the general state-of-the-art of the field, this section aims to provide an integrated and structured analysis of all works in the literature together from the viewpoint of the following three aspects:

\begin{enumerate}
    \item The databases, datasets, or repositories used for developing relevant methods or tools;
    \item The application goals, which usually belong to one or more subphases of the DbA process, based on a four-phase analogy model in the cognitive science field (depicted in Fig.~\ref{fig:fig4}).
    \item The AI or data science algorithms used in the data-driven DbA studies.
\end{enumerate}

\begin{figure}
	\centering
	\includegraphics[width=16cm]{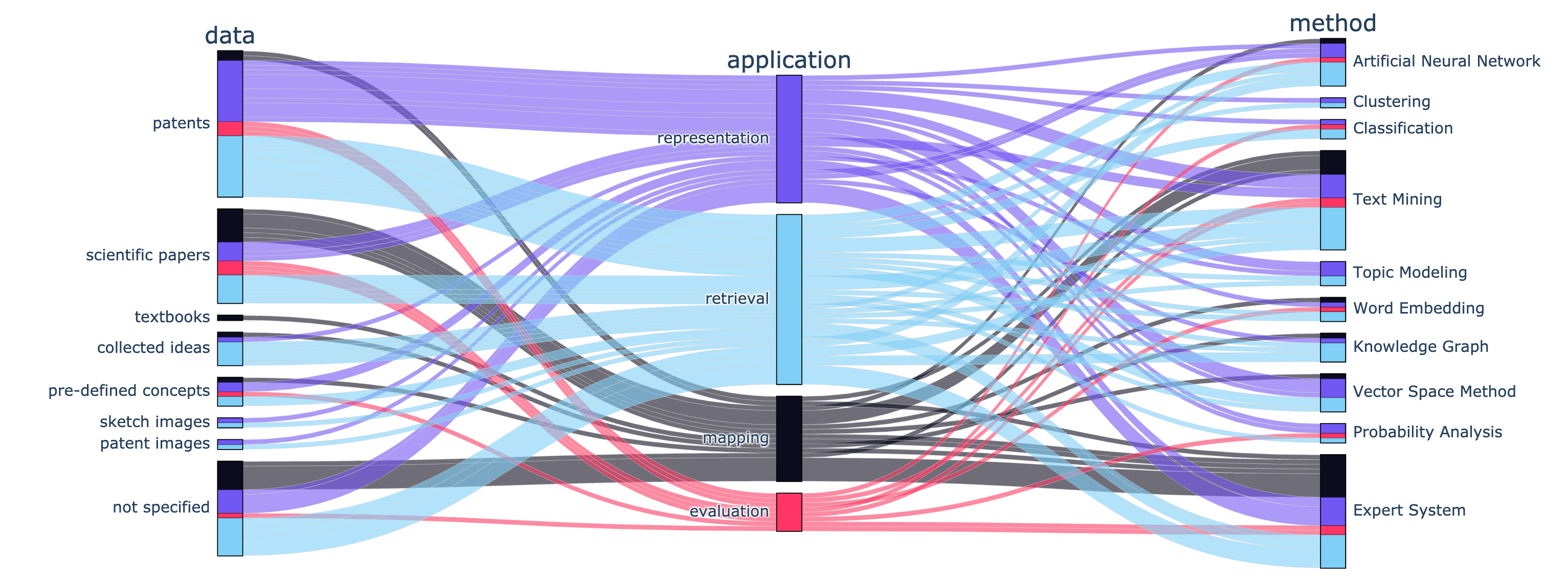}
	\caption{Overview of the state-of-the-art for data-driven DbA studies}
	\label{fig:fig5}
\end{figure}

Fig.~\ref{fig:fig5} presents a visual summary of the datasets, methods, applications of existing DbA studies, as well as the coupling of the three dimensions. A single DbA method or tool may contain more than one DbA subphase. We separately counted each item in Fig.~\ref{fig:fig5}, ensuring that each used algorithm or method was matched to the corresponding applications. The different line colors denote different phases in the DbA process. We further developed a web-based interactive data visualization\footnote{https://ddi.sutd.edu.sg/data-driven-design-by-analogy/} for public users to probe such coupling relationships among the three perspectives in individual prior studies.

\subsection{Data}
Table~\ref{tab:table1} summarizes the datasets used in DbA, including databases specifically developed for DbA research. The domain-specific knowledge databases, including DANE and Idea-Inspire, were manually created based on human-coding structures. The representation of design and relevant concepts by using predefined models is a time-consuming and labor-intensive task and requires domain expertise. Consequently, these repositories are usually small scale and lack scalability. For example, in the DANE repository, building a complete FBS model of a biological system requires 40–100 hours \cite{vattam2011dane}; thus, the repository contains only 40 cases and 22 complete FBS models. These models lack a uniform description and generalizability to enable processing large-scale databases, which are pre-conditions for adopting modern AI algorithms. While scholars have proposed building large and scalable digital libraries of analogical design case studies for broader usages \cite{goel2015on}, how to represent, aggregate and organize existing scattered and unstructured design data efficiently with a consistent structure remains an open question.

Compared with human-curated databases, patent and scientific paper databases are naturally structured and continually populated over time and cover a wide range of domains \cite{sarica2020technet,Shi2017}. They are public big-data sources that enable the development of scalable data-driven DbA methods and tools. Specifically, patent documents contain rich engineering design information of the structures, functions, mechanisms, and principles that might be useful for the design inspiration process \cite{murphy2014function,Jiang2021,Luo2021,srinivasan2018does,fu2013discovering,Singh2009inno}.

\begin{center}
\begin{table}
	\caption{Existing databases that have been adopted or developed for DbA}
	\centering
	\begin{tabular}{p{2cm}<{\centering}p{6cm}p{7cm}}
		\toprule
		\textbf{Category}    & \textbf{Data}    & \textbf{Size and refs} \\
		
		\midrule
		
		\multirow{3}{4em}{Patents} & Texts from the whole USPTO database & More than 6 million patent documents \cite{Luo2021,sarica2020technet}\\
		\cline{2-3}
        & Texts from the subset of the USPTO database & 155,000\cite{Verhaegen2011,verhaegen2011effectiveness,vandevenne2016seabird}, 6,500 \cite{murphy2014function}, 10,400 \cite{Song2020Anexp}, 200 \cite{song2017patent}, 100 \cite{fu2013discovering} patent texts\\ 
        \cline{2-3}
        & Images from the subset of the USPTO database & Images from the subset of the USPTO database \cite{Jiang2021} \\ 
        
        \midrule
        
        \multirow{4}{4em}{Scientific paper/books} & Papers from computer science conferences & Papers from computer science conferences \\
		\cline{2-3}
        & Papers from biological journals & 500 \cite{giler2014exploring}, 8,000 \cite{vandevenne2016seabird}, 10,000 \cite{kruiper-etal-2020-laymans} papers \\
        \cline{2-3}
        & Papers from engineering journals & More than 1 million papers \cite{Shi2017} \\ 
        \cline{2-3}
        & A biological textbook & A biological textbook \cite{cheong2011bio} \\
        
        \midrule
        
        \multirow{2}{4em}{Collected ideas} & Concepts collected from Quirky platform & The corpus cover over 8,500 products \cite{gilon2018analogy} \\
		\cline{2-3}
        & Data crawled from the Internet & Unknown \cite{han2018combinator,kwon2019visual} \\
        
        \midrule
        
        Sketches & Quickdraw & 75,000 sketches \cite{zhang2020unsupervised} \\
        
        \midrule
         
        \multirow{4}{4em}{Biomimicry databases} & AskNature & Over 1,700 biological concepts \cite{deldin2014asknature}\\
		\cline{2-3}
        & DANE & 40 FBS-based biological systems \cite{vattam2011dane} \\
        \cline{2-3}
        & Idea-Inspire & Over 100 biological and engineering SAPPhIRE-based concepts \cite{siddharth2018evaluating} \\ 
        \cline{2-3}
        & FOBIE & Over 1,500 annotated sentences with ~4,700 relations \cite{kruiper-etal-2020-laymans} \\
		
		\midrule
         
        \multirow{3}{4em}{Semantic Network-based databases} & WordNet & ~155,000 entities with ~648,000 relations \cite{miller1995wordnet} \\
		\cline{2-3}
        & ConceptNet & ~517,000 entities with ~$1.3\times{10^{11}}$ relations \cite{speer2017conceptnet} \\
        \cline{2-3}
        & TechNet & Over 4 million technical terms with their relationships \cite{sarica2020technet} \\
		
		\bottomrule
	\end{tabular}
	\label{tab:table1}
\end{table}
\end{center}

Recently, extensive research has established that crowdsourced design solution data collected from online ideation and task-orientated platforms (e.g., InnoCentive, Quirky, OpenIDEO, and Amazon Mechanical Turk) are useful for DbA research \cite{Kittur2019,goucher2019crowdsourcing,he2019mining}. The crowdsourced solutions make it possible to mine the real-world ideation outputs of complex cognitive tasks generated by thousands of people across the world. In addition, some researchers have observed that crowdfunding platforms (e.g., Kickstarter and Indiegogo) also accumulate product design projects and contain rich design information about product designs, and are therefore potentially useful data sources for analogical design stimuli \cite{song2020crowdfunding}.

Majority of the existing DbA research only focused on textual data and ignored data in other modalities such as visual information. Recently, some scholars have explored the mining of the large-scale 2D design image datasets (such as patent images and sketches) using various computer vision techniques. Other 2D image datasets and 3D model datasets, such as Shape-Net \cite{chang2015shapenet}, the Thingiverse repository\footnote{https://www.thingiverse.com/}, the IKEA design dataset \cite{lim2013parsing}, and mechanical CAD model dataset \cite{koch2019abc}, can be explored and utilized to develop tools for supporting DbA with visual analogies or multimodal analogies.

Furthermore, there is a paucity of gold-standard benchmarking datasets developed for specific DbA tasks. Consequently, the comparison and evaluation of different methods and tools in terms of their utilities for the same task have been plagued by ambiguities. Researchers of this community are recommended to collaborate on creating large-scale and canonically acceptable benchmark evaluation datasets for various situations and tasks in the DbA process. For example, the recently released FOBIE dataset \cite{kruiper-etal-2020-laymans} was developed to support the semi-open relationship extraction for nature-inspired engineering. Such a benchmarking dataset enables the measurement of the effectiveness of novel algorithms against those of other state-of-the-art approaches.

\subsection{Applications}
Based on the four-phase model of making analogies in cognitive science, we assessed the applications of the DbA tools and methods, and categorized each of them into one or more phases accordingly. From Table~\ref{tab:table2}, it can be distinctly observed that most of them fall within the analogy representation and retrieval subphases, which are the foundations of the entire DbA process. In most studies on these two subphases, text mining-related techniques were adopted to represent and identify candidate analogies for further use. Some researchers have utilized modern deep learning-based AI techniques to construct high-dimensional latent space for design data representation in different forms. These researchers generally posit that, guided by the identified stimuli, human users can continue the mapping and inference process. Although many AI-based methods and tools can recommend the candidate analogy list for users, few can interpret the results such as identifying the type of similarities.

Some studies sought to automate the analogical mapping process, including some systems that predate the 2000s, most of which are rule-based expert systems, as shown in Fig.~\ref{fig:fig5}. Some pre-defined rules, such as syntactic grammar-based rules \cite{cheong2014retrieving}, are embedded within these tools to facilitate automated inference. In addition, some tools implement idea generation by simply combining two ideas within moderate distance for design ideation \cite{han2018combinator}. Overall, current analogy mapping methods and tools have employed little machine intelligence, which is the major advantage of modern AI techniques. For example, recent deep learning AI models, such as the Siamese network and its variations \cite{Rossiello2019}, enable machines to learn pair-wise entity to entity embeddings that encode the specific type of relational information. If we collect a large set of analogy pairs, we may leverage these models to learn the analogical relational representation and conduct the analogical reasoning process.

Regarding the fourth phase, only a few researchers have attempted to evaluate the analogies or ideas generated by the DbA. This may be attributed to, at least, two reasons. First, it is difficult to establish the evaluation criteria. The usefulness and novelty that define “creativeness” may vary according to specific personal needs, organizational contexts, regional cultures, and times \cite{Boden1998}. Second, analogy assessment is a domain-related process. A traditional idea or concept in one knowledge domain might be novel in another. Therefore, the space that is newly transformed through analogy reasoning may require novel types of evaluation. A powerful evaluation mechanism would determine the effectiveness and usefulness of new data-driven DbA systems. There is still much room for researchers to develop robust, interpretable, and convincing analogy evaluation methods.

\begin{center}
\begin{table}
    \caption[Caption for LOF]{The applications of DbA tools and methods\protect\footnotemark}
	\centering
	\begin{tabular}{p{6cm}<{\centering}p{2cm}<{\centering}p{2cm}<{\centering}p{2cm}<{\centering}p{2cm}<{\centering}}
		\toprule
		\textbf{Tools or Methods}    & \textbf{Analogical Encoding}    & \textbf{Analogical Retrieval} & \textbf{Analogical Mapping} & \textbf{Analogical Evaluation}  \\
		
		\midrule
		
		TRIZ, 1956 \cite{al1999}             &   &   & * &   \\
		ARGO, 1988 \cite{huhns1988argo}             & * & * &   &   \\
		IDEAL, 1996 \cite{bhatta1996design}            & * & * & * &   \\
		KRITIK, 1997 \cite{goel1997kritik}           & * & * & * &   \\
		BIO-TRIZ, 2002 \cite{vincent2002systematic}         &   &   & * &   \\
		Pat-Analyzer, 2007 \cite{Cascini2007}     &   &   & * &   \\
		E2B, 2010 \cite{nagel2010engineering}, 2011 \cite{cheong2011bio}   &   &   & * &   \\
		PAnDA, 2011 \cite{verhaegen2011effectiveness}            & * & * &   &   \\
		WordTree, 2012 \cite{linsey2012design}          & * & * &   &   \\
		DANE, 2012 \cite{vattam2011dane}             & * & * &   &   \\
		IEKG, 2012 \cite{reich2012interdisciplinary}             &   &   & * &   \\
		Fu et al., 2013 \cite{fu2013discovering}        & * &   &   &   \\
		Func. Vector, 2014 \cite{murphy2014function}      & * & *\cite{fu2015design}  &   &   \\
		Four-box, 2014 \cite{helms2014thefour}         &   &   &   & * \\
		Cheong et al., 2014 \cite{cheong2014retrieving}    &   & * & * &   \\
		Glier et al., 2014 \cite{giler2014exploring}     &   &   &   & * \\
		AskNature, 2014 \cite{deldin2014asknature}        & * &   &   &   \\
		DRACULA, 2015 \cite{lucero2015design}          & * & * & * &   \\
		SEABIRD, 2016 \cite{vandevenne2016seabird}          & * & * & * &   \\
		Analogy Finder, 2016 \cite{Mccaffrey2016}   &   & * &   &   \\
		Sanaei et al., 2017 \cite{Sanaei2017}    &   & * &   &   \\
		Song et al., 2017 \cite{song2017patent}     &   & * &   &   \\
		B-Link, 2017 \cite{Shi2017}           & * & * &   & *\cite{han2020data} \\
		Gilon et al., 2018 \cite{gilon2018analogy}     &   & * &   &   \\
		SOLVENT, 2018 \cite{Chan2018}          &   & * & * &   \\
		Retriever, 2018 \cite{han2018computational}        &   & * &   &   \\
		Combinator, 2018 \cite{han2018combinator}       &   &   & * &   \\
		Idea-Inspire, 2018 \cite{siddharth2018evaluating}     & * & * & * & * \\
		Kwon et al., 2019 \cite{kwon2019visual}      &   & * &   &   \\
		TechNet, 2020 \cite{sarica2020technet}          & *\cite{Sarica2021a} & *\cite{Sarica2021idea} &   & *\cite{han2020data} \\
		VISION, 2020 \cite{Song2020Anexp}            & * & * &   &   \\
		Zhang et al., 2020 \cite{zhang2020unsupervised}     & * & * &   &   \\
		FOBIE, 2020 \cite{kruiper-etal-2020-scientific}            &   &   & * &   \\
		Jiang et al., 2021 \cite{Jiang2021}     & * & * &   &   \\
		InnoGPS, 2021 \cite{Luo2021}          & * &   & * &   \\
		
		\bottomrule
		
	\end{tabular}
	\label{tab:table2}
\end{table}
\end{center}

\subsection{Methods}
The AI-based techniques that have already been used in DbA studies are summarized in Table~\ref{tab:table3}. As shown in Fig.~\ref{fig:fig5}, the broadly defined NLP methods, including text mining, topic modeling, semantic networks, and pre-trained word embeddings, have played essential roles in supporting all stages of DbA. These methods aim at understanding and mining the textual nuances of natural language. However, some of them tackled the rough source data using only traditional textual pre-processing methods. They developed some syntactic rules for the processed dataset, which were still limited to the human experience. These pre-defined rules cannot be updated automatically to match the rapid advances in AI technologies. Some researchers that adopted knowledge graph techniques generally relied on the commonsense semantic networks trained on non-engineering data sources such as WordNet and ConceptNet. However, engineering designers have perceptions of technical terms that are distinct from the commonsense knowledge \cite{sarica2020technet}. In this case, the technological and engineering semantic networks (e.g., TechNet and B-link) that were recently developed based on engineering data can serve as infrastructure to support broad DbA studies.

\footnotetext{The star symbol "*" indicates the method/tool itself involved in a specific application. The reference after the star symbol indicates that the method/tool has been utilized in another study for a specific application goal.}

\begin{center}
\begin{table}
	\caption{Existing AI-based methods and algorithms that have been adopted or developed for DbA}
	\centering
	\begin{tabular}{p{3cm}p{8cm}p{4cm}}
		\toprule
		\textbf{Approach}    & \textbf{Method}    & \textbf{Refs} \\
		
		\midrule
		\multirow{3}{3cm}{Artificial Neural Network (ANN)} & Convolutional Neural Network (CNN) & \cite{Jiang2021,zhang2020unsupervised}\\
		\cline{2-3}
        & Recurrent Neural Network (RNN) & \cite{sarica2020technet}\\ 
        \cline{2-3}
        & Bi-directional Recurrent Neural Network (Bi-RNN) & \cite{gilon2018analogy,kruiper-etal-2020-scientific}\\ 
        \cline{2-3}
        & Long short-term memory (LSTM) & \cite{Sanaei2017}\\
        
        \midrule
		\multirow{3}{3cm}{Clustering} & Agglomerative Hierarchical Clustering (AHC) & \cite{song2017patent}\\
		\cline{2-3}
        & Relevance score-based clustering (RSC) & \cite{Song2020Anexp,fu2013discovering}\\
        
        \midrule
		\multirow{3}{3cm}{Classification} & Support Vector Machine (SVM) & \cite{Sanaei2017,giler2014exploring}\\
		\cline{2-3}
        & Naive Bayes (NB) & \cite{giler2014exploring}\\
        \cline{2-3}
        & K-Nearest Neighbors (KNN) & \cite{giler2014exploring}\\
		
		\midrule
		\multirow{3}{3cm}{Text mining} & Text preprocessing techniques & \cite{siddharth2018evaluating,sarica2020technet,Shi2017,gilon2018analogy,Chan2018,Cascini2007,Mccaffrey2016,cheong2014retrieving,verhaegen2011effectiveness,vandevenne2016seabird} \\
		\cline{2-3}
        & Syntactic Analysis & \cite{cheong2014retrieving,kruiper-etal-2020-scientific}\\
        
        \midrule
		\multirow{3}{3cm}{Word Embedding} & Word2Vec & \cite{siddharth2018evaluating,sarica2020technet,Chan2018}\\
		\cline{2-3}
        & Doc2Vec & \cite{Chan2018}\\
        \cline{2-3}
        & Glove & \cite{Chan2018}\\
		\cline{2-3}
        & ELMO & \cite{kruiper-etal-2020-scientific}\\
        
        \midrule
		\multirow{3}{3cm}{Topic-modeling} & Latent Semantic Analysis (LSA) & \cite{fu2013discovering}\\
		\cline{2-3}
        & Latent Dirichlet Allocation (LDA) & \cite{goucher2020adaptive}\\
        \cline{2-3}
        & Non-Negative Matrix Factorization (NMF) & \cite{Song2020Anexp}\\
        
        \midrule
		\multirow{3}{3cm}{Semantic Network and Knowledge Graph} & WordNet & \cite{linsey2012design,Sanaei2017}\\
		\cline{2-3}
        & ConceptNet & \cite{han2018combinator,han2018computational}\\
        \cline{2-3}
        & Cyc & \cite{gilon2018analogy}\\
		\cline{2-3}
        & TechNet & \cite{Sarica2021a,han2020data,Sarica2021idea}\\
        \cline{2-3}
        & B-Link & \cite{han2020data}\\
        
        \midrule
		\multirow{3}{3cm}{Vector Space Method (VSM)} & Functional vector space & \cite{murphy2014function,fu2015design}\\
		\cline{2-3}
        & Technology space map & \cite{Luo2021}\\
        \cline{2-3}
        & PA and OA space & \cite{verhaegen2011effectiveness,vandevenne2016seabird}\\
        
        \midrule
		\multirow{3}{3cm}{Probability analysis} & Dijkstra’s shortest path algorithm & \cite{Shi2017}\\
		\cline{2-3}
        & Bayesian inference approach & \cite{fu2013discovering}\\
        
        \midrule
        Expert system & Heuristic rules-based strategy & \cite{siddharth2018evaluating,goel1997kritik,bhatta1996design,huhns1988argo,reich2012interdisciplinary,al1999,vincent2002systematic,cheong2011bio,nagel2010engineering,vattam2011dane,helms2014thefour,lucero2015design,kwon2019visual}\\
        
		\bottomrule
	\end{tabular}
	\label{tab:table3}
\end{table}
\end{center}

Regarding the other methods, clustering algorithms have been used to represent design data by categories and facilitate the retrieval process \cite{Song2020Anexp,song2017patent,fu2013discovering}. Vector Space Method represents the original data as a vector of subitems, which also benefits both the encoding and retrieval subphases \cite{murphy2014function,Luo2021,fu2015design,verhaegen2011effectiveness,vandevenne2016seabird}. It is worth mentioning that recent deep learning techniques based on various structure models have shown great potential in some human tasks such as object recognition and classification. Specifically, in DbA studies, some initial intents have been employed to use CNNs and RNNs to support DbA tasks, ranging from analogy representation to analogy mapping \cite{Jiang2021,sarica2020technet,Sanaei2017,gilon2018analogy,kruiper-etal-2020-scientific,zhang2020unsupervised}.

The rapid advances in deep learning-based AI techniques may provide many powerful tools for data-driven DbA. For example, visual and semantic analogy question solving has been explored extensively \cite{Rossiello2019,Sadeghi2015deepv,Lu2019,Sadeghi2015visualogy,liao2017visual}; majority of the studies involve discovering an extendable mapping from an image and word pair and applying it to another image and word pair to find analogies. For example, Sadeghi et al. \cite{Sadeghi2015visualogy} utilized a quadruple Siamese network architecture to address the problem associated with creating visual analogies for natural images. Similarly, Rossiello et al. \cite{Rossiello2019} used hierarchical Siamese networks to learn relational representations by textual analogy. Moreover, various deep learning models, besides deep RNNs and CNNs and their variations, generative adversarial networks (GAN) \cite{goodfellow2014gan}, transformer \cite{vaswani2017attention} and graph neural networks (GNN) \cite{wu2020comprehensive}, have proven effective for learning complex features from datasets. More recently, the transformer-based models built by Open.AI, such as BERT \cite{devlin2018bert}, GPT-3 \cite{brown2020language}, and DALL·E \cite{Ramesh2020}, have exhibited record-breaking performances in understanding and generating new texts and images from given conditions. DbA researchers may adopt these state-of-the-art models from the frontier of AI research to support relational analogy representation and analogical reasoning for engineering design. However, note that the model interpretability of deep learning has always been a limiting factor for real-world use cases, where model outputs should be explained. Developing explainable AI (XAI) models \cite{arrieta2020explainable} for DbA tasks is still challenging for the community currently and needs more exploration.

\section{Future Opportunities and Directions}
In this paper, we have analyzed the literature on data-driven DbA to elucidate its state-of-the-art. Although existing methods and tools have already shown their ability to support the DbA process, many areas remain under-researched. In this section, we map the course of feasible directions for future research from the perspective of data, methods, applications, and their interactions.

\textbf{(1) New data sources exploration}

Based on our literature review, patent and scientific paper databases are suitable data sources because of their large size, as well as the rich information they possess on potential design inspiration. Both have been used in many data-driven DbA studies \cite{murphy2014function,Luo2021,Chan2018}. Although crowdsourced design data \cite{Kittur2019,goucher2019crowdsourcing,he2019mining} and crowdfunded project data \cite{song2020crowdfunding} have been explored in design research and are also potentially useful for DbA studies, they have not been fully exploited. In addition, most of the current data-driven DbA methods were designed  to merely mine textual information as analogy candidates; only a few researchers have focused on the data of other modalities \cite{Jiang2021,zhang2020unsupervised}. Scholars have revealed that designers often prefer visual representations to textual descriptions for idea generation \cite{linsey2011experimental}. Therefore, various kinds of visual design sources, such as technical images \cite{Jiang2021}, design sketches \cite{zhang2020unsupervised}, 3D product models \cite{chang2015shapenet,lim2013parsing,koch2019abc}, ought to be exploited and mined. It is recommended that researchers develop more tools based on the new data sources and broad data modalities.

\textbf{(2) Benchmark datasets construction}

The lack of distinct ground-truth and gold-standard benchmarks in DbA-related problems makes it impossible to directly compare different DbA methods and evaluate their performance in DbA applications reliably. For example, different analogy retrieval methods are typically applied to different datasets; literature \cite{gilon2018analogy} used a scientific paper database and literature \cite{Verhaegen2011} used a patent database. In this case, authors often define their own criteria for assessing performance in the absence of ground truth, which makes it difficult to determine the relative performance, strengths, and weaknesses of each method. This also prevents the community from following, adopting, and iterating on the state-of-the-art algorithms. Summers et al. \cite{Summers2017} similarly discussed the necessity and benefits of establishing benchmark datasets for function representation in engineering design. As a broader field, data-driven DbA requires a series of benchmark datasets for four different subprocesses. Some attempts have been made to create such benchmarking standard, such as the recently released FOBIE dataset \cite{kruiper-etal-2020-laymans} that was developed to support the semi-open relationship extraction for nature-inspired engineering. However, in practice, we should note that the ground truth labels used as the benchmarking standard may suffer from the same biases as the training data. We recommend that researchers create a series of benchmark datasets by combining different data sources for DbA tasks to surmount such limitations.

\textbf{(3) Engineering knowledge graph-based applications}

Data-driven DbA requires the knowledge database in the backend. When it comes to the large and cross-domain ones, the knowledge base needs to be structured and organized to enable effective retrieval and mapping. In the engineering design literature, the network map of technology domains \cite{srinivasan2018does,song2017patent} and semantic networks \cite{sarica2020technet,Shi2017} have been created and used to guide the retrieval of analogical design stimuli based on the quantified knowledge distance between domains \cite{Luo2021} or the semantic distance between concepts \cite{Sarica2021idea}. Meanwhile, the most frequently used knowledge bases are the large pre-trained commonsense knowledge graphs such as WordNet, ConceptNet and FreeBase \cite{Han2021,camburn2020machine}.

In the field of AI, the knowledge graph notion specifically refers to the qualitative networks composed of many heterogenous <entity, relation, entity> triplets \cite{zaveri2016quality,paulheim2017knowledge}. The relations are qualitatively specific and sensible for humans, in contrast to the one-dimensional semantic distance or knowledge distance constructs. The past decade has seen the rapid adoption of large knowledge graphs in powering knowledge representation, retrieval, and mapping tasks in many AI applications. Google, Facebook, Amazon, TikTok, among others, have developed and used knowledge graphs to support the search, recommendation, and query engines in their AI products and services. 

Therefore, it is recommended that researchers extend the uses of such large pre-trained cross-domain semantic networks (e.g., B-Link \cite{Shi2017} and TechNet \cite{sarica2020technet}) or knowledge graphs (e.g., WordNet \cite{miller1995wordnet} and ConceptNet \cite{speer2017conceptnet}) as knowledge bases for developing knowledge-based expert systems for DbA. Furthermore, it will be a promising effort to develop graph knowledge bases in which the entities are presented at the hierarchical domain to concept levels and both the quantitative and qualitative relations between entities are pre-trained, to augment analogy retrieval and mapping across domains and concepts. 

\textbf{(4) Deep learning solutions, new models, architectures, and XAI}

Deep learning techniques based on various structure models, such as CNNs and RNNs, have shown their strengths in supporting DbA tasks, ranging from analogy representation to analogy mapping \cite{Jiang2021,sarica2020technet,gilon2018analogy,kruiper-etal-2020-scientific,zhang2020unsupervised,Kittur2019}. On the basis of these efforts, the up-to-date basic deep learning structures (e.g., transformer \cite{vaswani2017attention} and GNN \cite{wu2020comprehensive}) and large-scale pre-trained models (e.g., BERT \cite{devlin2018bert} and GPT \cite{brown2020language}) ought to be explored and utilized to facilitate all DbA subprocesses. There exist many opportunities for collaborative research among the psychological cognition, engineering design, and deep learning communities, and new models and algorithms can be developed to handle various types of data sources. Specifically, in analogy mapping and evaluation tasks, domain expertise is required to interpret the results yielded by deep learning algorithms. 

In addition, understanding the fundamental properties and interpretability of deep learning is an essential part of such data-driven studies. For example, when we use deep learning to retrieve analogy candidates, being aware of the similarity type of analogies is as important as obtaining the recommended analogy candidate list. The retrieval systems should provide users more insightful information with the final outputs. In this regard, one of the pioneer works is SOLVENT \cite{Chan2018}, which attempts to interpret the types of analogies by annotating and mapping different aspects of research papers. Beyond this, researchers may combine the modern deep learning with traditional interpretable machine learning techniques (e.g., decision trees and Bayesian classifiers) to develop more explainable models for DbA.

\textbf{(5) New subprocess of data-driven DbA: Analogy-based design synthesis}

In the traditional DbA process, following the analogy retrieval and mapping steps, the final design synthesis stage typically requires the experience, intelligence, and efforts of designers. As a comparison, various large-scale datasets collected and represented for previous DbA subprocesses also provide opportunities for scholars to explore the deep learning-based data synthesis and generative methods, which we term as a new subprocess of data-driven DbA, analogy-based design synthesis. Unlike traditional design synthesis strategies, such as shape grammars and constraint programming, data-driven design synthesis methods do not necessarily require expert knowledge and can automatically learn to generate plausible new designs from datasets \cite{chen2021padgan,Chen2019synthesizing,oh2019deep,Nobari2021,yang2018microstructural,shu20203d,gan2021ijie}. 

Among various data-driven design synthesis methods, recent deep generative models have gained significant attention because of their ability to learn complex features from samples. GAN \cite{goodfellow2014gan} and Variational autoencoder (VAE) \cite{rezende2014stochastic} are the most commonly used deep generative models to assist engineering design. Specifically, variations of GAN have been used in designing airfoils \cite{chen2021padgan,Chen2019synthesizing}, car wheels \cite{oh2019deep}, bicycles \cite{Nobari2021}, airplane \cite{shu20203d} and social robots \cite{gan2021ijie}, and VAE has been used in designing material microstructures \cite{yang2018microstructural}. Recently, the pre-trained GPT \cite{brown2020language} and DALL·E \cite{Ramesh2020} models released by Open.AI have exhibited record-breaking performances in creating novel texts and images. As these synthesis methods typically require given inputs as the initial design information, the identified analogies from the previous DbA steps may potentially guide the direction of design generation. Therefore, it is recommended that researchers extend the use of these existing AI-based generative models and algorithms to develop analogy-based design synthesis tools. 

\begin{figure}
	\centering
	\includegraphics[width=15cm]{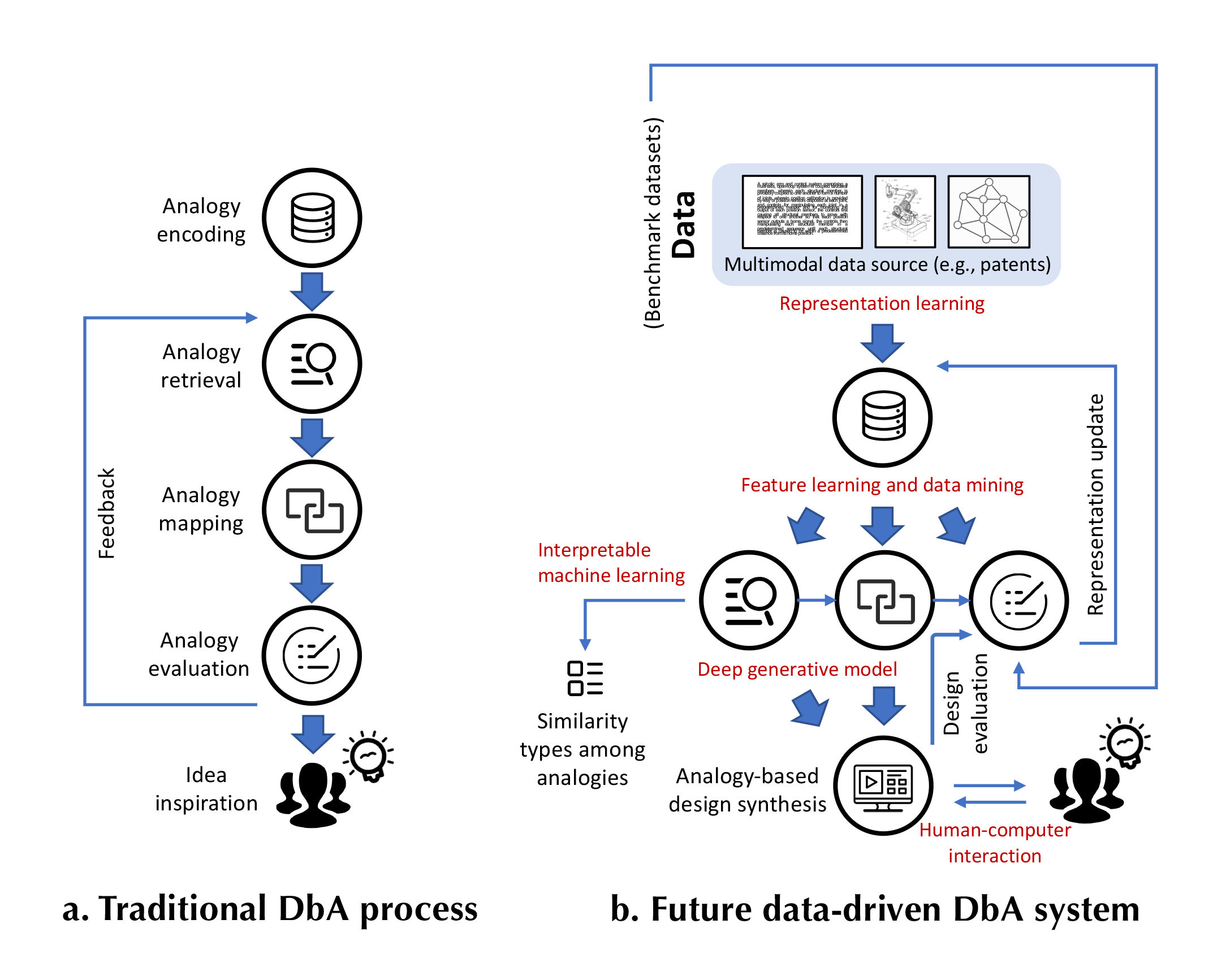}
	\caption{Schematic comparison between the traditional DbA process and future data-driven DbA system}
	\label{fig:fig6}
\end{figure}

Taken together, the foregoing propositions can be merged and implemented in one future data-driven DbA system that applies various latest AI techniques to different DbA subprocesses. In Fig.~\ref{fig:fig6}, we conceptualized such a data-driven DbA system, in comparison with the traditional one. In this future system, multi-source and multimodal design-related data are embedded and stored in a high-dimensional vector space through representation learning. The vector space is dynamic and can be updated based on the requirements and performance of the downstream analogical tasks. Analogical retrieval, mapping and evaluation will be conducted based on the distributed representation of design data samples through various data mining techniques. Benchmark datasets are available for researchers to make comparisons among new and alternative algorithms. During analogical reasoning, interpretable machine learning will enable us peek into the support evidence of analogy making, such as providing similarity types among analogies. Finally, deep generative models will be utilized to automatically generate new designs according to the recommended analogy pairs and their mapping relationship. In this system, human-computer interaction may play an essential role in this process to improve the quality of design synthesis. These newly created designs can also be assessed by the intelligent evaluation model that considers user needs, sentiments, and compassion. 

Furthermore, it is important to note that the design of future data-driven intelligent DbA systems should be grounded on the scientific understanding of analogy. The available cognitive science foundations of analogy had been mainly developed via traditional and small human experiments and remained abstract and at a macro-level of understanding and application \cite{christensen2007relationship,markman1993structural,markman2009supporting,Gentner2013,hey2008analogies,gill2019dimensions,tsoka2020dimensions,linsey2008modality,fu2014bioinspired}. With the growing development and uses of data-driven DbA systems, we anticipate a growing amount of data on various DbA processes and outcomes to be generated. Such big data may offer new research opportunities by analyzing them to nuance and advance our scientific understanding of the cognitive process and mechanisms of analogy. In turn, the new understanding can be applied to develop more explainable AI systems for design by analogy.

\section{Conclusion}
Data-driven Design-by-Analogy presents great potential to augment design creativity and enhance innovation. We contribute to the literature by elucidating the state-of-the-art data-driven DbA studies and identifying new and feasible future research opportunities. Our structured review of the data, applications, and methods of existing data-driven DbA systems and tools provides a foundation or basis for researchers to follow the relevant latest efforts in a coherent form. Several research directions are proposed in terms of data, applications, and methods, respectively, to bridge the gap between engineering design and AI research. We hope that our review and propositions may guide the efforts of researchers aiming to develop more powerful and intelligent data-driven DbA methods, tools, and systems.

In this paper, although we only focus on data-driven DbA methods and tools, AI and data science techniques can be also useful to augment other classical engineering design methods \cite{Chiarello2021}, such as TRIZ \cite{al1999,Cascini2007}, design heuristics \cite{yilmaz2016design,jin2021design}, design principles \cite{Singh2009inno,fu2016designprinciple}, design structure matrix \cite{clarkson2004predicting,eppinger2012design}, product family and platform design \cite{simpson2014advances,song2019platform}, first principles \cite{merrill2002first,cagan1991dimensional,kannengiesser2018ekphrasis}, C-K \cite{hatchuel2018design}, blending \cite{nagai2009concept} and combinational design \cite{he2019mining,He2019thesis}. We hope that researchers of these relevant fields may also find inspirations from this paper to advance the data-driven approaches for their design methodologies with latest AI and data science technologies.

\section*{Acknowledgements}
The authors acknowledge the funding support for this work received from the SUTD-MIT International Design Center and SUTD Data-Driven Innovation Laboratory (DDI, https://ddi.sutd.edu.sg/), National Natural Science Foundation of China (52035007, 51975360), Special Program for Innovation Method of the Ministry of Science and Technology, China (2018IM020100), National Social Science Foundation of China (17ZDA020). The support from CU Denver’s College of Engineering, Design and Computing (CEDC, https://engineering.ucdenver.edu/) and the Comcast Media and Technology Center (CMTC, https://comcastmediatechcenter.org/) are also acknowledged. Any ideas, results and conclusions contained in this work are those of the authors, and do not reflect the views of the sponsors.

\bibliographystyle{unsrtnat}
\bibliography{references}  






\end{document}